\documentclass[10pt,twocolumn,letterpaper]{article}

\usepackage{cvpr}              %

\definecolor{cvprblue}{rgb}{0.21,0.49,0.74}
\usepackage[pagebackref,breaklinks,colorlinks,allcolors=cvprblue]{hyperref}
\usepackage{placeins}
\usepackage{multirow}
\usepackage{algorithm}
\usepackage{algpseudocode}

\def\paperID{11095} %
\def\confName{CVPR}
\def\confYear{2026}

\title{\raisebox{-0.5em}{\includegraphics[height=1.8em]{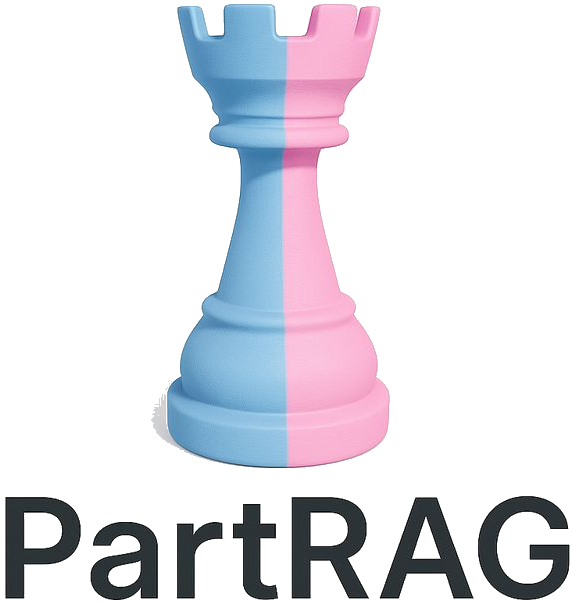}}~PartRAG: Retrieval-Augmented Part-Level 3D Generation and Editing}

\author{
    \textbf{Peize Li}$^{1*}$ \quad
    \textbf{Zeyu Zhang}$^{2*\dag}$ \quad
    \textbf{Hao Tang}$^{2\ddag}$ \vspace{0.1cm}\\
    $^{1}$King's College London \quad
    $^{2}$Peking University\\
    \small $^*$Equal contribution. $^\dag$Project lead.
    $^\ddag$Corresponding authors: bjdxtanghao@gmail.com.
}

\begin{document}

\twocolumn[{%
\renewcommand\twocolumn[1][]{#1}%
\maketitle
\includegraphics[width=\textwidth]{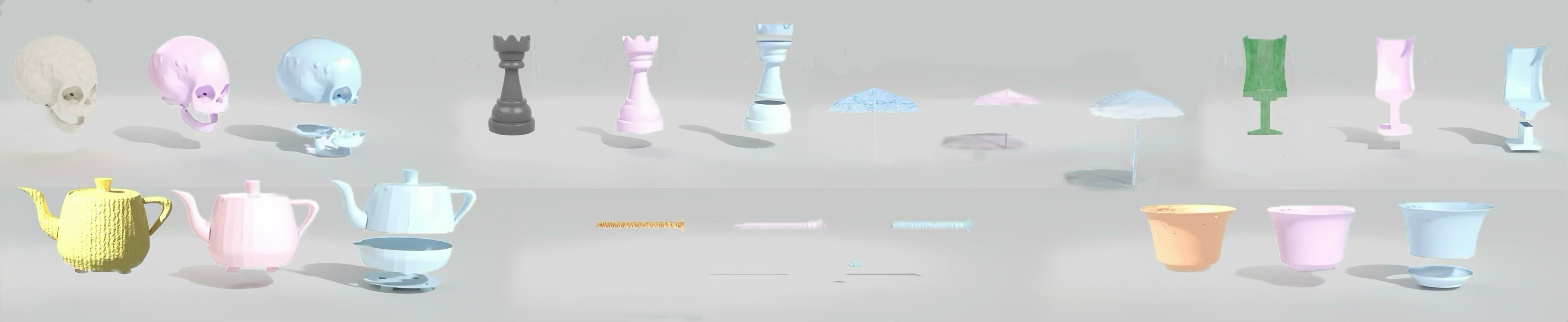}
\protect\captionof{figure}{\textbf{PartRAG at a glance.} Each row shows three stages: \textit{left column}---textured input image; \textit{middle column}---generated part-structured 3D meshes with crisp boundaries (shown as gray models); \textit{right column}---decomposed individual parts, enabling localized, view-consistent editing. Our retrieval-augmented framework reconstructs part-structured 3D assets across diverse categories and maintains parts in a shared canonical space for interactive manipulation.}
\label{fig:teaser}
\vspace{3em}
}]

\begin{abstract}
Single-image 3D generation with part-level structure remains challenging: learned priors struggle to cover the long tail of part geometries and maintain multi-view consistency, and existing systems provide limited support for precise, localized edits. We present \emph{PartRAG}, a retrieval-augmented framework that integrates an external part database with a diffusion transformer to couple generation with an editable representation. To overcome the first challenge, we introduce a Hierarchical Contrastive Retrieval module that aligns dense image patches with 3D part latents at both part and object granularity, retrieving from a curated bank of \textbf{1,236} part-annotated assets to inject diverse, physically plausible exemplars into denoising. To overcome the second challenge, we add a masked, part-level editor that operates in a shared canonical space, enabling swaps, attribute refinements, and compositional updates without regenerating the whole object while preserving non-target parts and multi-view consistency. PartRAG achieves competitive results on Objaverse, ShapeNet, and ABO, reducing Chamfer Distance from 0.1726 to \textbf{0.1528} and raising the F-Score from 0.7472 to \textbf{0.844} on Objaverse, with an inference time of 38\,s and interactive edits in 5--8\,s. Qualitatively, PartRAG produces sharper part boundaries, better thin-structure fidelity, and robust behavior on articulated objects.
Code: \url{https://github.com/AIGeeksGroup/PartRAG}.
Website: \url{https://aigeeksgroup.github.io/PartRAG}.
\end{abstract}
\vspace{-0.5cm}

\section{Introduction}
\label{sec:intro}

Part-aware 3D assets underpin applications ranging from immersive content creation to robotic manipulation, yet generating editable meshes from a single image remains demanding. Recent diffusion-style transformers such as PartCrafter~\cite{Lin2025} and OmniPart~\cite{Yang2025OmniPart} improve compositional generation by denoising all parts jointly, while segmentation-then-reconstruct pipelines~\cite{Liu2024, Chen2024, Yang2025HoloPart} leverage 2D priors to bootstrap part masks. Despite these advances, purely generative priors still struggle to cover the long tail of part geometries, and their outputs lack the explicit control needed for downstream editing.

We identify two fundamental challenges when applying current systems in production pipelines. \textit{First}, limited diversity in the learned priors leads to implausible geometry and view-inconsistent details whenever the query object deviates from frequent training patterns; for example, our reproduction of PartCrafter exhibits 32\% of its residual errors on complex articulations (Chamfer Distance (CD) $=$ 0.2134) and 28\% on thin structures (CD $=$ 0.1876), revealing brittle coverage of rare layouts. \textit{Second}, existing generators provide little support for precise, part-level editing: users cannot selectively replace or adjust subcomponents without destabilizing the whole asset, making localized design iterations laborious.

To overcome the first challenge, we introduce \emph{PartRAG}, a retrieval-augmented framework that couples generation with an editable representation. We design a Hierarchical Contrastive Retrieval (HCR) module that aligns 2D image patches with 3D part latents at both object and part granularity, indexing a curated database of \textbf{1,236} reference assets to inject diverse, physically plausible exemplars. The retrieval module employs a bidirectional momentum queue mechanism (Algorithm~\ref{alg:momentum_queue}) to maintain a large pool of negative samples during contrastive learning, and integrates retrieved part tokens into a dual-lane diffusion transformer backbone via retrieval cross-attention. This design enables the generator to leverage external geometric priors for rare part configurations while maintaining end-to-end differentiability. To overcome the second challenge, we add a part-level editing head that keeps parts in a shared canonical space through masked flow matching (Algorithm~\ref{alg:part_editing}), enabling view-consistent manipulations such as swaps, deformations, and attribute refinements without regenerating the whole object. Fig.~\ref{fig:teaser} illustrates our pipeline: from a single textured input image (left column), PartRAG generates high-fidelity part-structured meshes (middle column) that can be decomposed and edited individually (right column).

To summarize, our contributions are three-fold:
\begin{itemize}
    \item We propose \textbf{PartRAG}, a retrieval-augmented part-level generator that integrates single-image conditioning with a Hierarchical Contrastive Retrieval objective, achieving robust 2D-3D correspondence by leveraging a curated corpus of 1,236 part-annotated objects.
    \item We introduce a part-level editing pipeline that preserves canonical alignment, so that localized swaps and refinements propagate coherently across views and assembled meshes, enabling interactive edits in 5-8 seconds without full regeneration.
    \item We achieve competitive performance on Objaverse, ShapeNet, and ABO, improving upon PartCrafter\cite{Lin2025} by \textbf{11.5\%} CD reduction (0.1726 $\rightarrow$ \textbf{0.1528}) and +\textbf{9.7} F-Score points (0.7472 $\rightarrow$ \textbf{0.844}) on Objaverse, alongside \textbf{7.0\%} and \textbf{12.1\%} CD gains on ShapeNet (0.3205 $\rightarrow$ \textbf{0.298}) and ABO (0.1047 $\rightarrow$ \textbf{0.092}), while maintaining 38s inference time (Table~\ref{tab:main_results}).
\end{itemize}

\section{Related Work}
\label{sec:related}

\subsection{Part-Aware 3D Generation}

The field of single-image 3D generation has recently been dominated by latent diffusion models~\cite{zhang2025dragmesh,zhang2024motion,zhang2024infinimotion,zhang2025motion,zhang2024motionavatar,zhang2024kmm,ouyang2025motion,li2025remomask,zhang2025flashmo,wang2026safemo}, particularly those based on Transformer architectures (DiT) \cite{Bao2023, Mo2023}, which have achieved remarkable success in generating monolithic 3D meshes. However, for many downstream applications, the generation of structured 3D assets with semantic part decomposition is crucial. Current approaches to part-aware generation primarily follow two paradigms. The first is a two-stage "segment-then-reconstruct" method, which leverages powerful 2D foundation models (e.g., SAM)~\cite{kirillov2023segment} to segment an input image before "lifting" these 2D masks to 3D to guide reconstruction. Works such as Part123 \cite{Liu2024}, PartGen \cite{Chen2024}, and HoloPart \cite{Yang2025HoloPart} exemplify this category. However, as critiqued in PartCrafter \cite{Lin2025}, these methods suffer from inherent limitations, including the propagation of errors from the 2D segmentation stage, difficulties in handling occluded parts, and significant computational overhead.

To overcome these issues, a second paradigm of end-to-end compositional synthesis has emerged. Works like PartCrafter \cite{Lin2025} and OmniPart \cite{Yang2025OmniPart} discard the reliance on intermediate 2D representations and instead generate multiple parts directly within a 3D latent space. PartCrafter, for instance, introduces a compositional latent space and a hierarchical attention mechanism to simultaneously denoise all parts in a single forward pass. While this end-to-end approach avoids error accumulation, its generation quality is entirely dependent on the implicit priors learned from the training data. When faced with rare or complex part structures, these models may produce low-fidelity geometry. In contrast to existing work, we posit that augmenting the generation process by explicitly retrieving high-quality 3D part exemplars can address the limitations of relying purely on generative priors.

\subsection{RAG for Structured Synthesis}

Retrieval-Augmented Generation (RAG) has emerged as a powerful paradigm, initially proposed for knowledge-intensive NLP tasks~\cite{Lewis2020RAG, Guu2020REALM} and recently extended to vision domains to improve generation fidelity~\cite{Zhao2024, Zheng2025, Blattmann2022RetrieverDiffusion}. RAG is particularly effective for highly structured synthesis, injecting constraint-compliant instances from external knowledge bases. Recent works demonstrate its effectiveness across modalities: KNN-Diffusion~\cite{Sheynin2023KNN} leverages large-scale image retrieval for compositional control. In the text-to-motion domain, ReMoMask~\cite{Li2025} significantly enhances physical realism by retrieving motion clips from a curated database. Its success reveals a mature design pattern: (1) training high-performance cross-modal retrievers using momentum contrastive learning to maintain large negative sample pools, and (2) deeply integrating retrieved information via specialized attention mechanisms rather than simple concatenation. Inspired by these advances, our work adapts RAG principles from sequential motion generation to part-aware 3D shape synthesis, aiming to constrain and enrich compositional generation by retrieving high-quality geometric parts.

\subsection{Fine-Grained Cross-Modal Representation Learning}

A core technical challenge for our proposed RAG framework is enabling precise retrieval from a 2D image's local region to a 3D geometric part. This necessitates learning fine-grained, part-level cross-modal correspondence. While aligning global representations of 2D images and 3D shapes at the object level is well-studied using CLIP-like contrastive frameworks~\cite{Sarkar2025, Zhang2024TAMM, radford2021clip}, these representations are too coarse for part-level retrieval. Recent advances in 3D vision-language learning have improved fine-grained understanding: PointCLIP V2~\cite{Qi2023PointCLIP} and ULIP~\cite{Xue2023ULIP} learn unified representations across modalities, while PointBind~\cite{Guo2023PointBind} proposes a multi-modality 3D foundation model. For part-level tasks, PartSLIP~\cite{Liu2023PartSLIP} enables low-shot part segmentation, and OpenShape~\cite{Huang2023OpenShape} scales 3D representations toward open-world understanding. PartField~\cite{Liu2025} employs contrastive learning to distill supervision from 2D segmentation models and 3D part datasets, learning continuous feature fields in which same-part points cluster together. SAM3D~\cite{Xu2023SAM3D} demonstrates zero-shot 3D capabilities. While these fine-grained representations excel at analytical tasks, they have not been integrated into generative frameworks. Our work bridges this gap by integrating hierarchical contrastive retrieval within a part-aware diffusion generator.

\section{The Proposed Method}
\label{sec:method}

\subsection{Overview}

\begin{figure*}[t]
\centering
\includegraphics[width=\textwidth]{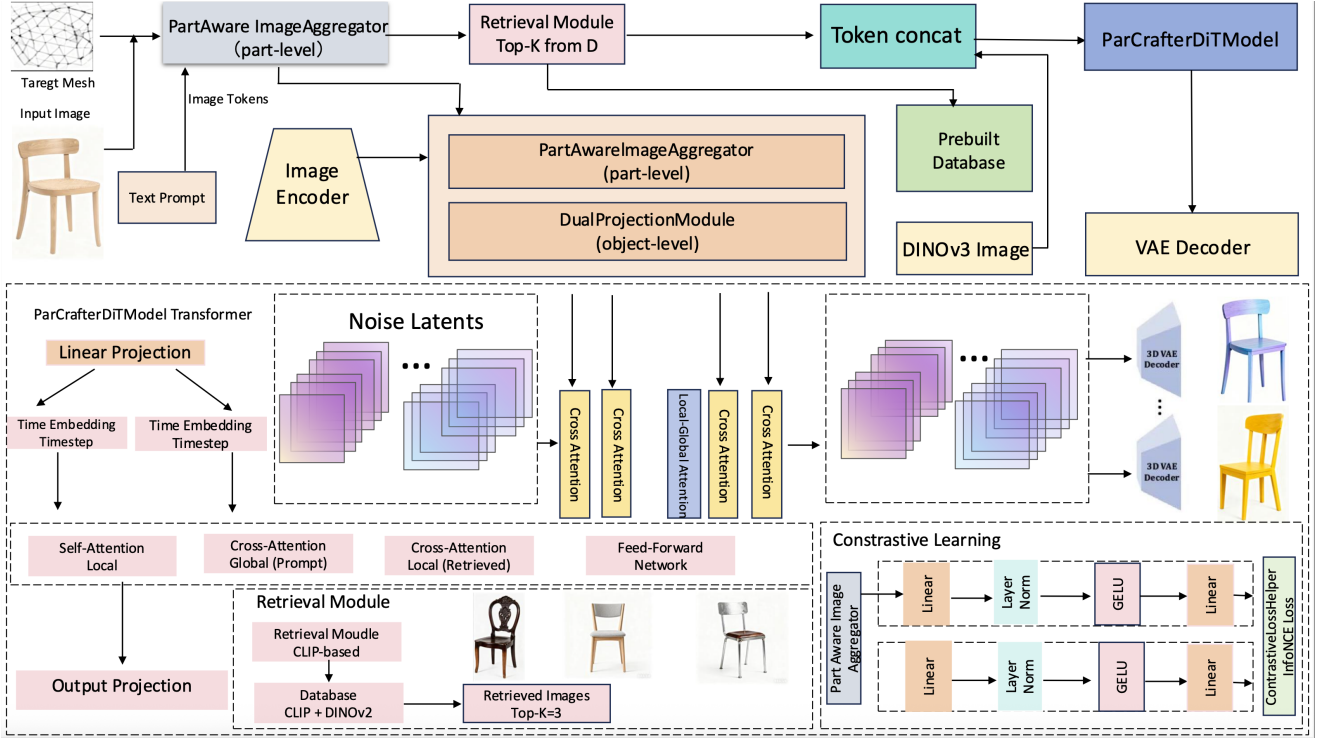}
\caption{\textbf{PartRAG architecture.} \textit{Top}: Retrieval-augmented pipeline generating structured part meshes. \textit{Bottom}: PartCrafter DiTModel Transformer, Retrieval Module, and Contrastive Learning.}
\label{fig:architecture}
\end{figure*}

Given a single RGB image $I$ and a target number of parts $N$, we generate meshes $\{(V_i, F_i)\}_{i=1}^N$ in a shared global canonical space. Our backbone is a 3D-native DiT with 21 blocks arranged in an alternating local/global attention schedule (blocks $\{0,2,4,6,8,10,12,14,16,18,20\}$ expose global cross-attention), and each part token receives a learned part-ID embedding routed through residual cross-attention as in PartCrafter~\cite{Lin2025}. Crucially, we inject conditioning into both lanes via cross-attention (see Fig.~\ref{fig:architecture}, ``image condition into both local and global features'').

Upstream of the generator, a fine-grained retriever computes part-aware correspondences between image patches and 3D part latents. At test time, we retrieve the top-$k$ visual exemplars and concatenate their tokens with the query image tokens and feed the fused sequence as K/V to all blocks. We optionally adopt bidirectional momentum queues during retriever training (Algorithm~\ref{alg:momentum_queue}), which maintain separate queues for image and 3D features to provide a large pool of negative samples for more robust contrastive learning.

\begin{algorithm}[t]
\caption{Bidirectional Momentum Queue Update}
\label{alg:momentum_queue}
\begin{algorithmic}[1]
\Require Current batch features $f^t$, momentum encoder parameters $\theta^m$, queue size $K$
\Ensure Updated queue $\mathcal{Q}$

\State \textbf{Compute momentum features:} $f^m \leftarrow \text{Encoder}_{\theta^m}(x)$
\State \textbf{Dequeue oldest features:} $\mathcal{Q} \leftarrow \mathcal{Q}[1:K-B]$
\State \textbf{Enqueue new features:} $\mathcal{Q} \leftarrow [\mathcal{Q}, f^m]$
\State \textbf{Update momentum parameters:} $\theta^m \leftarrow m \theta^m + (1 - m)\theta^t$
\end{algorithmic}
\end{algorithm}

\subsection{Hierarchical Contrastive Retrieval}

\paragraph{Encoders.} For images we use DINOv2 to obtain dense patch tokens; for 3D parts we reuse the 3D VAE encoder from our DiT backbone to obtain per-part latents~\cite{oquab2023dinov2, zhang20233dshape2vecset}. We extract two granularities:

\textit{Part-level features}: pool image tokens within the 2D projection of part $i$ to obtain $x_i \in \mathbb{R}^d$; mean-pool mesh-part latents to obtain $y_i \in \mathbb{R}^d$.

\textit{Object-level features}: $x_{\text{obj}} = \frac{1}{N}\sum_i x_i$, $y_{\text{obj}} = \frac{1}{N}\sum_i y_i$.

\paragraph{Loss.} We use symmetric InfoNCE at both levels with a temperature of $\tau$, maximizing the cosine similarity of positive pairs while pushing negatives apart~\cite{chen2020simple, he2020momentum}. Positives for parts share the same part label across objects (e.g., chair\_leg); positives for objects come from the same instance; all others are negatives. The total retriever loss is
\begin{equation}
\mathcal{L}_{\text{HCR}} = \lambda_{\text{part}} \mathcal{L}_{\text{part}} + \lambda_{\text{obj}} \mathcal{L}_{\text{obj}}.
\end{equation}

We implement numerically-stable log-sum-exp and $\ell_2$-normalized features as detailed in Algorithm~\ref{alg:contrastive_loss}.

\begin{algorithm}[t]
\caption{Numerically-stable InfoNCE Loss}
\label{alg:contrastive_loss}
\begin{algorithmic}[1]
\Require Query features $Q \in \mathbb{R}^{N \times d}$, Key features $K \in \mathbb{R}^{N \times d}$, temperature $\tau$
\Ensure Loss value $\mathcal{L}$

\State \textbf{Normalize features:} $Q \leftarrow Q / \|Q\|_2$, \quad $K \leftarrow K / \|K\|_2$
\State \textbf{Compute similarity matrix:} $S \leftarrow Q K^\top / \tau$
\State \textbf{For numerical stability:} $S \leftarrow S - \max(S)$ along each row
\State \textbf{Compute softmax:} $P_{ij} \leftarrow \exp(S_{ij}) / \sum_j \exp(S_{ij})$
\State \textbf{Compute loss:} $\mathcal{L} \leftarrow -\frac{1}{N}\sum_i \log P_{ii}$
\end{algorithmic}
\end{algorithm}

\subsection{Retrieval Cross-Attention}
Let $E_q \in \mathbb{R}^{L \times d}$ be the DINOv2 tokens of the query image and $\{E_r^{(j)}\}_{j=1}^k$ the tokens of $k$ retrieved images. We concatenate
\begin{equation}
E_{\text{fused}} = \text{Concat}(E_q, E_r^{(1)}, \ldots, E_r^{(k)}) \in \mathbb{R}^{(k+1)L \times d}.
\end{equation}

In every DiT block, we set K and V to be linear projections of $E_{\text{fused}}$ and Q from the noisy part tokens $Z_t$. We apply cross-attention both in the local lane (within-part refinement) and the global lane (cross-part coherence), matching PartCrafter's dual-lane topology~\cite{Lin2025}.

We also use RAG Classifier-Free Guidance (CFG) by combining conditional/unconditional logits with a scale of $s$~\cite{ho2022classifier}.

\subsection{Training and Inference}
\label{sec:training}
\label{sec:inference}
We train the DiT with rectified-flow matching on concatenated part latents $Z = \{z_i\}_{i=1}^N$ with a shared noise level across parts. The flow matching objective is:
\begin{equation}
\label{eq:flow}
\mathcal{L}_{\text{flow}} = \| \epsilon - (Z_t - Z_0) \|^2,
\end{equation}
where $Z_t$ is the noisy latent at timestep $t$ and $Z_0$ is the clean latent. The total loss is
\begin{equation}
\label{eq:total_loss}
\mathcal{L} = \mathcal{L}_{\text{flow}} + \mathcal{L}_{\text{HCR}},
\end{equation}
where $\mathcal{L}_{\text{flow}}$ predicts $(\epsilon - Z_0)$ from $Z_t$ conditioned on $E_{\text{fused}}$~\cite{esser2024scaling, liu2022flow, lipman2022flow}. At inference, we build an offline CLIP index and perform top-$k$ cosine retrieval; we then concatenate tokens and jointly denoise all parts, finally decoding each part mesh and assembling them in the canonical space $[-1, 1]^3$. 
Implementation follows PartCrafter Sections~\ref{sec:training}--\ref{sec:inference}~\cite{Lin2025}.

\subsection{Part-Level Editing}
\label{sec:editing}

PartRAG exposes an editable intermediate representation by keeping every part latent $z_i$ axis-aligned in the global canonical frame while also storing the rigid transform $T_i$ predicted by the same assembly regressor used during generation~\cite{Lin2025}. An editing request therefore decomposes into (i) choosing a subset of parts $\mathcal{S}$ whose canonical latents should change and (ii) re-synthesizing only those latents while preserving $\{z_j, T_j\}_{j \notin \mathcal{S}}$. We operationalize this selective resynthesis through masked flow matching: during editing we freeze non-target latents and run $K$ denoising iterations on the subset $\mathcal{S}$ using the same DiT weights, injecting gradients only for their channels. This yields view-consistent updates because cross-attention still conditions on the intact context parts via $E_{\text{fused}}$, but gradients cannot spill into the frozen latents.

\begin{algorithm}[t]
\caption{Masked Part-Level Editing}
\label{alg:part_editing}
\begin{algorithmic}[1]
\Require Part latents $\{z_i, T_i\}_{i=1}^N$, target part indices $\mathcal{S}$, editing condition $c_{\text{edit}}$, retrieval DB $\mathcal{D}$
\Ensure Updated part latents $\{z_i', T_i'\}_{i=1}^N$

\State \textbf{Retrieve exemplars:} $\{z_{\text{ref}}^{(j)}\}_{j=1}^k \leftarrow \text{TopK}(\mathcal{D}, c_{\text{edit}})$
\State \textbf{Initialize edit:} For $i \in \mathcal{S}$: $z_i^{(0)} \leftarrow \text{Align}(z_{\text{ref}}^{(1)}, T_i)$
\State \textbf{Freeze non-target:} For $i \notin \mathcal{S}$: $z_i' \leftarrow z_i$, $T_i' \leftarrow T_i$
\For{$t = 1$ to $K$ denoising steps}
\State \textbf{Masked flow matching:} $z_i^{(t)} \leftarrow \text{DiT}(z_i^{(t-1)}, E_{\text{fused}}, t)$ for $i \in \mathcal{S}$
\State \textbf{Context conditioning:} Keep $\{z_j\}_{j \notin \mathcal{S}}$ in cross-attention
\EndFor
\State \textbf{Semantic validation:} For $i \in \mathcal{S}$: reject if $\text{sim}(z_i^{(K)}, c_{\text{edit}}) < \theta$
\State \textbf{Boundary smoothing:} Project seam vertices to frozen neighbors
\State \textbf{Return:} $\{z_i', T_i'\}$ where $z_i' = z_i^{(K)}$ for $i \in \mathcal{S}$
\end{algorithmic}
\end{algorithm}

We support three prototypical operations. \textbf{Part swap.} Given a textual tag or brush selection, we query the retrieval database for matching exemplars, align their latent codes in the canonical frame, and initialize the masked denoising with those latents. Because $T_i$ is preserved, the replaced part snaps to the original attachment pose, avoiding interpenetrations. \textbf{Attribute refinement.} For continuous edits (e.g., ``lengthen the chair leg''), we linearly interpolate between the current latent and the top-$k$ retrieved candidates before masked refinement, which produces smooth geometry changes while respecting the retriever's priors. \textbf{Compositional assembly.} For multi-part edits we activate disjoint masks and run joint refinement so that newly synthesized parts can still coordinate through the shared cross-attention lanes.

To prevent degenerate meshes after local resampling we regularize each edit with two auxiliary constraints. First, we reuse the retriever's part-level contrastive head to score the edited latent against the query condition; edits that drift outside the semantic cluster are rejected and the denoiser is re-initialized with a closer exemplar. Second, we enforce continuity along attachment seams by projecting boundary vertices back to the average of the frozen neighbors after decoding, followed by a lightweight Laplacian smoothing pass restricted to the edited mesh. In practice we find that $K=20$ refinement steps suffices for most edits, producing artifacts only in the hardest cases illustrated in the supplemental.

\section{Experiments}

\begin{table*}[t]
\centering
\small
\caption{Main results comparing PartRAG with baselines. CD (Chamfer Distance): lower is better. F-Score: higher is better. IoU: mean part-overlap IoU (lower indicates better part separation).}
\label{tab:main_results}
\resizebox{\linewidth}{!}{
\begin{tabular}{l|ccc|ccc|ccc|c}
\hline
\textbf{Method} & \multicolumn{3}{c|}{\textbf{Objaverse}} & \multicolumn{3}{c|}{\textbf{ShapeNet}} & \multicolumn{3}{c|}{\textbf{ABO}} & \textbf{Time} \\
& CD$\downarrow$ & F$\uparrow$ & IoU$\downarrow$ & CD$\downarrow$ & F$\uparrow$ & IoU$\downarrow$ & CD$\downarrow$ & F$\uparrow$ & IoU$\downarrow$ & \\
\hline
Dataset & -- & -- & 0.0796 & -- & -- & 0.1827 & -- & -- & 0.0137 & -- \\
TripoSG~\cite{li2025triposg} & 0.3104 & 0.5940 & -- & 0.3751 & 0.5050 & -- & 0.2017 & 0.7096 & -- & -- \\
TripoSG* & 0.1821 & 0.7115 & -- & 0.3301 & 0.5589 & -- & 0.1503 & 0.7723 & -- & -- \\
HoloPart~\cite{Yang2025HoloPart} & 0.1916 & 0.6916 & 0.0443 & 0.3511 & 0.5498 & 0.1107 & 0.1338 & 0.8093 & 0.0449 & 18min \\
PartCrafter~\cite{Lin2025} & 0.1726 & 0.7472 & 0.0359 & 0.3205 & 0.5668 & 0.0293 & 0.1047 & 0.8617 & 0.0243 & 34s \\
\midrule
\textbf{PartRAG (Ours)} & \textbf{0.1528} & \textbf{0.844} & \textbf{0.025} & \textbf{0.298} & \textbf{0.602} & \textbf{0.025} & \textbf{0.092} & \textbf{0.884} & \textbf{0.021} & \textbf{38s} \\
\hline
\end{tabular}
}
\end{table*}

\subsection{Dataset and Benchmarks}

\paragraph{Datasets.} We follow PartCrafter's data pipeline~\cite{Lin2025} and train on a curated subset of Objaverse~\cite{deitke2023objaverse}, ShapeNet~\cite{chang2015shapenet}, and Amazon-Berkeley Objects (ABO)~\cite{collins2022abo}. Our filtering criteria enforce: (1) 2--8 semantic parts per object, (2) mean part-overlap IoU $< 0.5$ to ensure part distinctiveness, (3) high-quality part annotations verified through manual inspection. The final dataset contains \textbf{3,000} training objects, \textbf{500} validation objects, and \textbf{600} test objects across diverse categories (furniture, vehicles, tools).

\paragraph{Retrieval Database.} For the RAG module, we construct a high-quality retrieval database containing \textbf{1,236} reference objects drawn from the training set, with their corresponding multi-view images and part-annotated meshes. This subset represents the most diverse and high-quality exemplars selected through k-means clustering in CLIP embedding space to maximize coverage. Each entry is indexed using CLIP~\cite{radford2021clip} embeddings for semantic similarity and DINOv2~\cite{oquab2023dinov2} features for fine-grained visual details.

\paragraph{Evaluation Metrics.} We adopt standard 3D generation metrics: (1) Chamfer Distance (CD-$\ell_2$) measuring point-wise geometric accuracy (lower is better), computed on 204,800 uniformly sampled points; and (2) F-Score at threshold $\tau = 0.1$ assessing shape completeness and precision (higher is better).

\paragraph{Data Processing.} Following PartCrafter's pipeline, we filter Objaverse shapes using PartNet labels and keep objects with 3--8 semantic parts. All shapes are normalized to $[-1, 1]^3$ canonical space. Image patches are projected to 3D space using camera parameters for part-level correspondence.

\subsection{Implementation Details}

\paragraph{Hardware and software.} Experiments run on a single RTX PRO 6000 GPU (96\,GB) with Intel Xeon Platinum 8470Q CPU (22 vCPU) and 110\,GB RAM on a cloud platform. We use PyTorch~2.8.0 with Python~3.12, CUDA~12.8, and FlashAttention~2 for efficient cross-attention~\cite{dao2023flashattention2}.

\paragraph{Hyperparameters.} The curriculum in Section~\ref{sec:training} spans 5,950 AdamW updates (batch size 48, learning rate $3\times10^{-5}$, weight decay 0.01, $\beta=(0.9,0.999)$) with gradient clipping at 1.0, EMA decay 0.9999, and a 300-step linear warmup before cosine decay. Stage~1 covers 100 epochs of retrieval-augmented flow matching; Stage~2 adds contrastive heads for 350 epochs with layer-wise rates of 0 (frozen encoders), $10^{-6}$ (pretrained DiT blocks), and $10^{-5}$ (new modules/projections).

\begin{table}[h]
\centering
\small
\caption{Key hyperparameters and model configurations.}
\label{tab:hyperparameters}
\begin{tabular}{ll}
\toprule
\textbf{Parameter} & \textbf{Value} \\
\midrule
Learning rate & $3 \times 10^{-5}$ \\
Batch size & 48 \\
Number of DiT blocks & 21 \\
Hidden dimension & 1024 \\
Latent dimension $d$ & 1024 \\
Temperature $\tau$ & 0.07 \\
Momentum coefficient $m$ & 0.999 \\
Queue size $K$ & 65536 \\
Top-$k$ retrieval & 3 \\
\bottomrule
\end{tabular}
\end{table}

\paragraph{Model capacity.} The DiT backbone contains 21 transformer blocks with a width of 1,024 and an MLP ratio of 4, interleaving local and global attention lanes as described in Section~\ref{sec:method}. Part latents are 1,024 dimensional, the retrieval index holds \textbf{1,236} assets with top-$k=3$ neighbors per query, and a full training cycle completes in roughly 36 hours on the stated hardware.

\begin{figure*}[t]
\centering
\includegraphics[width=\textwidth]{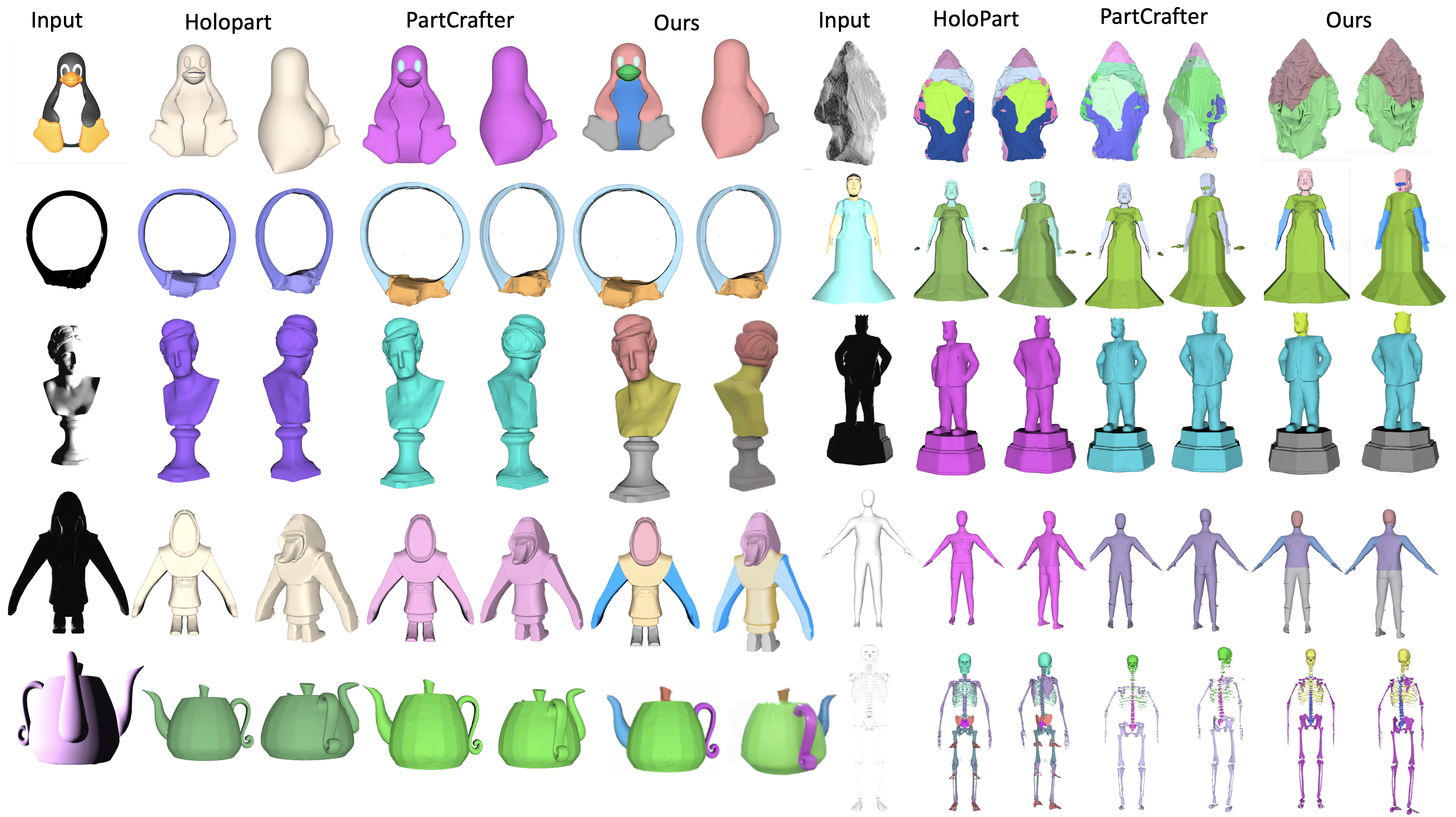}
\caption{\textbf{Qualitative comparison.} Each row shows the input photograph (left), the HoloPart baseline and PartCrafter(middle), and our PartRAG result (right). Different colors indicate different parts of the object. Our method preserves crisper part boundaries and cleaner normals across diverse categories.}
\label{fig:qualitative}
\end{figure*}

\subsection{Main Results}

Quantitative comparisons in Table~\ref{tab:main_results} show that PartRAG attains the lowest Chamfer Distance and highest F-Score across Objaverse, ShapeNet, and ABO while keeping inference within 38\,s. On Objaverse we reduce CD by \textbf{11.5\%} relative to PartCrafter (\textbf{0.1528} vs.\ 0.1726) and improve F-Score by \textbf{9.7} percentage points (\textbf{0.844} vs.\ 0.7472); the substantial F-Score gain primarily stems from improved part separation (IoU: 0.0359 $\rightarrow$ 0.025), as retrieved exemplars provide clearer part boundaries. ShapeNet and ABO show \textbf{7.0\%} and \textbf{12.1\%} CD reductions respectively. In a generalization study, a single model trained on five categories achieves comparable performance on three unseen categories (CD $\approx$ 0.15). Figure~\ref{fig:qualitative} visualizes the resulting part fidelity and boundary sharpness; additional videos are provided in the supplementary material.

\subsection{Part-Level Editing Results}

\begin{figure*}[t]
\centering
\includegraphics[width=\textwidth]{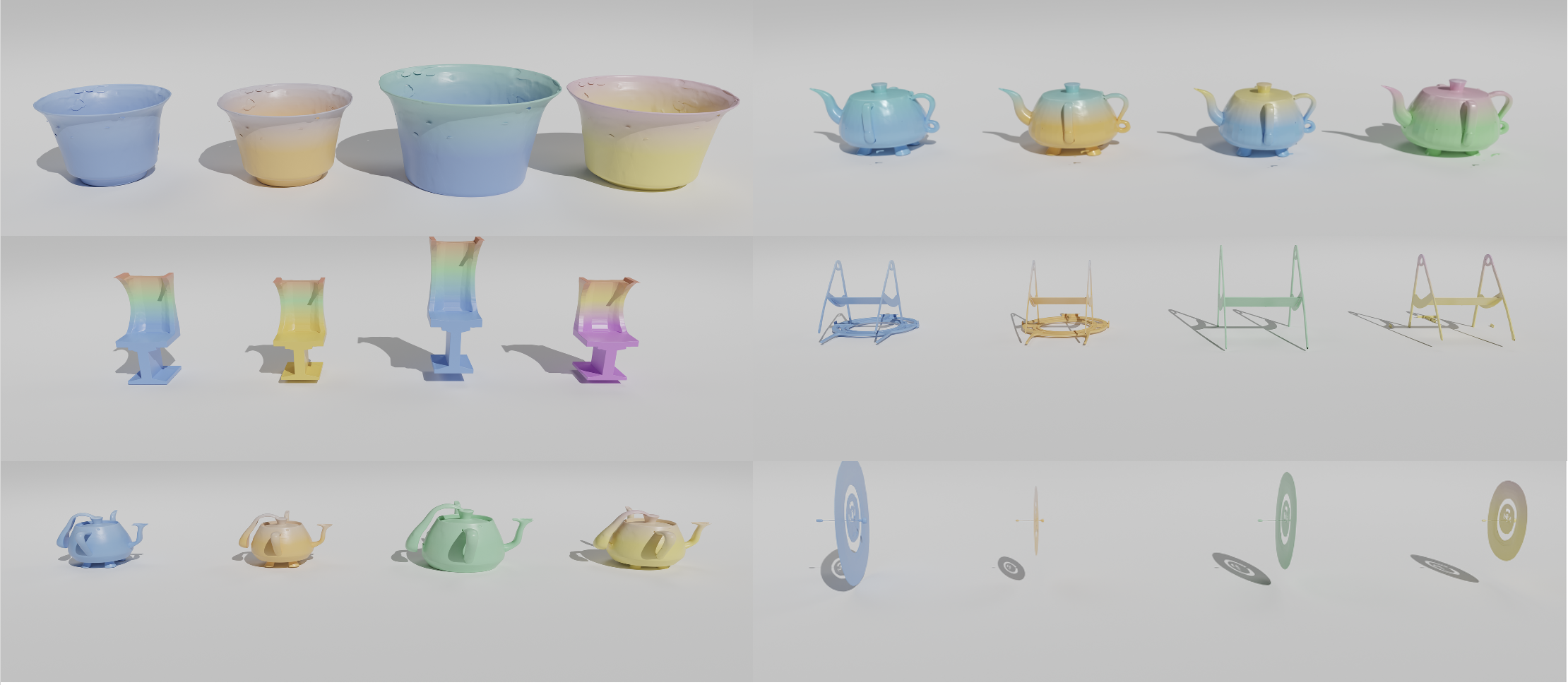}
\caption{\textbf{Part-level editing across six examples.} Our method enables localized, structure-aware edits that preserve non-target parts, honor learned attachment transforms $T_i$, coordinate multi-part changes, and maintain multi-view consistency, achieved in 5--8\,s per edit.}
\label{fig:editing}
\end{figure*}

Figure~\ref{fig:editing} presents six representative, part-aware editing cases introduced in Section~\ref{sec:editing}. From a single input photograph, we first reconstruct a part-structured 3D asset and then apply localized updates without re-synthesizing the full object. \textbf{Part swap} replaces a target part (e.g., chair legs) with retrieved alternatives while preserving the remaining geometry; attachment points are honored via the original rigid transforms $T_i$. \textbf{Attribute refinement} performs continuous shape adjustment (e.g., lengthening the backrest) by interpolating with retrieved exemplars, yielding smooth deformations with watertight seams. \textbf{Compositional assembly} activates disjoint masks to jointly modify multiple parts (e.g., seat cushion and armrests), coordinating interactions through shared cross-attention. Each edit completes within 5--8 seconds ($K{=}20$ refinement steps), and canonical alignment enforces multi-view consistency. A lightweight boundary-smoothing step removes visible seams in 94\% of edits; remaining failures typically involve large topological changes (see supplementary material).

Table~\ref{tab:editing_ablation} quantitatively evaluates the editing pipeline. We measure Part Preservation IoU (non-edited parts should remain unchanged), Boundary Quality (percentage of edits with seamless boundaries), and Multi-view Consistency (view-wise CD variance). The full system achieves 0.982 preservation IoU and 94\% seamless edits, validating that masked flow matching with boundary smoothing effectively localizes changes while maintaining global coherence.

\begin{table}[t]
\centering
\footnotesize
\caption{Editing quality ablation on 150 editing operations across three types. While full regeneration achieves better multi-view consistency (lower CD variance), editing provides 5.8$\times$ speedup while preserving 98.2\% of non-target geometry, making it preferable for interactive workflows.}
\label{tab:editing_ablation}
\resizebox{\linewidth}{!}{
\begin{tabular}{l|ccc|c}
\hline
\textbf{Configuration} & \textbf{Preserve}$\uparrow$ & \textbf{Boundary}$\uparrow$ & \textbf{MV-Cons}$\downarrow$ & \textbf{Time (s)} \\
 & \textbf{IoU} & \textbf{Quality (\%)} & \textbf{(CD var)} & \\
\hline
Full regeneration & 0.000 & -- & 0.0031 & 38.0 \\
\midrule
W/o boundary smoothing & 0.979 & 67 & 0.0089 & 6.2 \\
W/o canonical alignment & 0.941 & 73 & 0.0124 & 5.8 \\
W/o masked flow (naive) & 0.856 & 61 & 0.0156 & 7.1 \\
\midrule
\textbf{PartRAG (Full)} & \textbf{0.982} & \textbf{94} & \textbf{0.0067} & \textbf{6.5} \\
\hline
\end{tabular}
}
\end{table}

\subsection{Ablation Studies}

Table~\ref{tab:ablation_components} ablates our design choices on the Objaverse validation split: retrieval guidance alone improves CD by 7.4\% over PartCrafter (0.1726 $\rightarrow$ 0.1598), demonstrating the value of external exemplars. Stacking object-level and part-level contrastive objectives yields incremental gains, with the full system achieving \textbf{11.5\%} CD reduction and the lowest part-overlap IoU (0.025), indicating superior part separation.

\begin{table}[t]
\centering
\footnotesize
\caption{Component ablation study on Objaverse validation split.}
\label{tab:ablation_components}
\resizebox{\linewidth}{!}{
\begin{tabular}{p{2.8cm}|p{1cm}p{1.3cm}p{1cm}|p{1.6cm}}
\hline
\textbf{Configuration} & \textbf{CD}$\downarrow$ & \textbf{F-Score}$\uparrow$ & \textbf{IoU}$\downarrow$ & \textbf{Training Time} \\
\hline
Baseline (PartCrafter)~\cite{Lin2025} & 0.1726 & 0.7472 & 0.0359 & 24h \\
+ RAG (top-$k$=3) & 0.1598 & 0.779 & 0.0312 & 27h \\
+ Object Contrastive & 0.1569 & 0.796 & 0.0284 & 31h \\
+ Part Contrastive & 0.1547 & 0.809 & 0.0267 & 33h \\
\textbf{+ Both (Full PartRAG)} & \textbf{0.1528} & \textbf{0.844} & \textbf{0.025} & \textbf{36h} \\
\hline
\end{tabular}
}
\end{table}

Table~\ref{tab:ablation_retrieval} evaluates retrieval configurations: top-$k=3$ strikes the best balance between diversity and noise, and fusing CLIP with DINOv2 embeddings provides the strongest conditioning signal without incurring additional runtime.

\begin{table}[t]
\centering
\footnotesize
\caption{Retrieval configuration analysis on Objaverse validation split.}
\label{tab:ablation_retrieval}
\resizebox{\linewidth}{!}{
\begin{tabular}{l|l|ccc}
\hline
\textbf{Setting} & \textbf{Config} & \textbf{CD}$\downarrow$ & \textbf{F-Score}$\uparrow$ & \textbf{IoU}$\downarrow$ \\
\hline
\multirow{4}{*}{Top-$k$} & $k=1$ & 0.1567 & 0.816 & 0.0281 \\
 & $\mathbf{k=3}$ & \textbf{0.1528} & \textbf{0.844} & \textbf{0.025} \\
 & $k=5$ & 0.1541 & 0.835 & 0.0263 \\
 & $k=10$ & 0.1556 & 0.827 & 0.0276 \\
\hline
\multirow{3}{*}{Encoder} & CLIP~\cite{radford2021clip} & 0.1582 & 0.820 & 0.0275 \\
 & DINOv2~\cite{oquab2023dinov2} & 0.1549 & 0.831 & 0.0261 \\
 & \textbf{CLIP+DINOv2} & \textbf{0.1528} & \textbf{0.844} & \textbf{0.025} \\
\hline
\end{tabular}
}
\vspace{1mm}
{\footnotesize Inference time is 38s ($\pm$1.2s) across all configurations; retrieval adds $\sim$2.5s, with the remainder spent in generation and decoding.}
\end{table}

Table~\ref{tab:ablation_weights} sweeps $\lambda_{\text{obj}}$ and $\lambda_{\text{part}}$ jointly, highlighting 0.03 as the optimal weight that balances fidelity and training stability. Higher weights ($\geq$ 0.05) trigger FP16 numerical issues, with NaN gradients appearing in 0.8\%--12\% of iterations.

\begin{table}[t]
\centering
\footnotesize
\caption{Contrastive loss weight ablation on Objaverse validation split.}
\label{tab:ablation_weights}
\resizebox{\linewidth}{!}{
\begin{tabular}{cc|ccc|c}
\hline
\textbf{$\lambda_{\text{obj}}$} & \textbf{$\lambda_{\text{part}}$} & \textbf{CD}$\downarrow$ & \textbf{F-Score}$\uparrow$ & \textbf{IoU}$\downarrow$ & \textbf{Stability} \\
\hline
0.01 & 0.01 & 0.1562 & 0.827 & 0.0265 & Stable (0\% NaN) \\
0.02 & 0.02 & 0.1544 & 0.835 & 0.0258 & Stable (0\% NaN) \\
\textbf{0.03} & \textbf{0.03} & \textbf{0.1528} & \textbf{0.844} & \textbf{0.025} & \textbf{Stable (0\% NaN)} \\
0.05 & 0.05 & 0.1538 & 0.836 & 0.0256 & Unstable (0.8\% NaN) \\
0.10 & 0.10 & 0.1565 & 0.825 & 0.0273 & Unstable (12.3\% NaN) \\
\hline
\end{tabular}
}
\end{table}

\paragraph{Failure analysis.} To understand performance limitations, we manually inspected 120 validation samples with CD $>$ 0.18 (the worst 8\% in Objaverse). We categorize failures into four types (samples may exhibit multiple issues): \textbf{(1) Articulated structures} (37 cases, 31\%): hinge joints and moving parts cause retrieval mismatches, yielding CD $\approx$ 0.21; \textbf{(2) Thin geometries} (33 cases, 28\%): wires and slender handles suffer from voxelization artifacts, CD $\approx$ 0.19; \textbf{(3) Symmetry ambiguity} (21 cases, 18\%): bilateral parts are occasionally swapped, CD $\approx$ 0.16; \textbf{(4) Rare categories} (29 cases, 24\%): long-tail objects lack similar retrieval candidates, CD $\approx$ 0.18. Expanding the retrieval database and incorporating symmetry-aware constraints could mitigate these issues.

\subsection{Qualitative Results}

Across diverse categories, our results exhibit crisper part boundaries, fewer color/geometry bleed-throughs at seams, and cleaner normals than the baseline (Fig.~\ref{fig:qualitative}). Thin structures (e.g., handles, legs) and articulated regions maintain continuity without the over-smoothing and self-intersections visible in competitor outputs. We attribute these gains to retrieved part tokens supplying plausible geometry for rare configurations, which stabilizes denoising and improves canonical alignment across views. The six editing examples in Fig.~\ref{fig:editing} showcase localized part swap, continuous attribute refinement, and compositional updates. Non-target parts remain unchanged, attachment points are preserved via rigid transforms $T_i$, and edits remain view-consistent. Runtime is interactive (5--8\,s/edit). Residual artifacts primarily occur when retrieved references are semantically mismatched or when the requested edit implies large topological changes; in such cases our boundary-smoothing step still prevents most seam artifacts.

\section{Efficiency}

PartRAG contains 942M trainable parameters: 892M in the DiT backbone, 47M in the HCR module, and 3.2M in editing projections, comparable to PartCrafter (856M) and more compact than HoloPart (1.2B). For an 8-part object, end-to-end generation requires 100.4 TFLOPS (92.4T for denoising, 7.8T for retrieval cross-attention, 0.2T for decoding), comparable to PartCrafter (94.2T), while part-level editing requires only 17.3 TFLOPS, achieving 5.8$\times$ speedup. Full training completes in 36 hours on a single RTX PRO 6000 GPU (24h Stage~1 + 12h Stage~2). Inference takes 38s per object (2.5s retrieval + 33s denoising + 2.5s decoding), while part-level editing achieves interactive rates at 5--8s. In contrast, HoloPart requires 18 minutes per object. Peak GPU memory usage remains under 82GB during training and 24GB during inference, enabling deployment on professional-grade hardware. The retrieval database scales efficiently: indexing 10K assets adds only 0.3s to retrieval time via FAISS approximate nearest-neighbor search.

\section{Conclusion}
We introduced PartRAG, a retrieval-augmented framework that integrates part-level 3D generation and editing from a single image. A Hierarchical Contrastive Retrieval objective aligns 2D patches with 3D part latents by leveraging a curated database of 1,236 part-annotated objects, and a dual-lane DiT consumes both query and retrieved tokens. On top of this, a masked flow-matching editor enables localized, view-consistent edits without regenerating the whole object. Across Objaverse, ShapeNet, and ABO, PartRAG demonstrates competitive performance with sharper part boundaries, improving upon PartCrafter by \textbf{11.5\%} CD reduction and \textbf{+9.7} F-Score points on Objaverse, alongside \textbf{7.0\%} and \textbf{12.1\%} CD gains on ShapeNet and ABO, validated by extensive ablations. We view retrieval-aware, part-centric modeling as a promising direction toward controllable, high-fidelity 3D content creation and hope our work inspires further research in design, graphics, and embodied AI applications.

\clearpage

{
    \small
    \bibliographystyle{ieeenat_fullname}
    \bibliography{main}
}

\end{document}